\documentclass[10pt,twocolumn,letterpaper]{article}

\usepackage{wacv}                            

\usepackage{pifont}

\usepackage{booktabs}

\definecolor{wacvblue}{rgb}{0.21,0.49,0.74}
\usepackage[pagebackref,breaklinks,colorlinks,allcolors=wacvblue]{hyperref}

\def\wacvPaperID{*****}
\def\confName{WACV}
\def\confYear{2027}

\title{What Do Interaction Representations Actually Measure?\\
Pre-Event Separability in Weakly-Supervised Violence Detection}

\author{Parishruthi Ganesh\\
Auburn University\\
{\tt\small pzg0050@auburn.edu}
}

\begin{document}
\maketitle

\begin{abstract}
Articulated human pose provides detailed body-configuration information beyond
coarse spatial relationships, but whether this detail yields greater
discriminative information when the downstream pipeline is held fixed remains
unclear. We examine this through early violence detection---where
distinguishing violent from benign close contact is the core
difficulty---holding the tracker, temporal
head, supervision, folds, and evaluation fixed, we compare five interaction
representations spanning coarse bounding-box geometry, a matched handcrafted
pose analogue, enriched pose descriptors, and a matched-capacity encoder
learned from raw joints, using video-level evaluation with cluster-bootstrap
confidence intervals. No pose-based representation outperforms coarse geometry, though with fifteen
anomalous videos this subset cannot rule out small effects.
Extending the same pipeline to frozen visual encoders, and repeating the geometry--appearance--context
comparison on XD-Violence ($137$ anomalous videos, nine times our UCF-Crime
anomalous sample), we find that person-crop appearance and whole-frame context both
exceed geometry by a large margin on both benchmarks, but that whole-frame
context matches appearance on UCF-Crime and \emph{exceeds} it on the
larger benchmark ($4/4$ seeds): cropping to the tracked interacting people
yields no advantage over encoding the entire frame, and on the larger split is
measurably worse than it. This
prompts a direct test of what the benchmark measures. Scoring anomalous videos using \emph{only} frames preceding the annotated event
onset, under a control that removes sequence length as a cue, retains
$39$--$91\%$ of above-chance separation across both benchmarks, including for
seven hand-designed geometric channels. Inspection of the tightest pre-onset windows identifies concrete provenance
artifacts: editorial title cards and platform watermarks absent from the
surveillance footage supplying the normal class.
Video-level AUC in this setting is thus a composite of event evidence and
pre-event source cues, the latter accounting for most of the probe separation
and a substantial fraction under the detection head---a large shared source of
discrimination that can obscure differences between representations. The diagnostic requires only the
temporal annotations these benchmarks already ship.
\end{abstract}


\section{Introduction}
\label{sec:intro}

Understanding how people interact is fundamental to surveillance understanding,
group-activity recognition, video anomaly detection, and violence detection.
Articulated human pose captures detailed body configurations beyond coarse
spatial relationships such as inter-person distance, and pose- and
skeleton-based models have shown strong performance in action understanding
\cite{yan2018stgcn,posec3d}. However, it remains unclear whether this added
representational detail provides greater discrimination than coarse geometry when
the downstream learning pipeline and model capacity are held fixed---whether
richer descriptions of an interaction carry more discriminative information, or
whether that detail is redundant with what simpler cues already capture.

This question, however, is rarely tested in isolation. Most interaction- and
pose-based approaches evaluate their representations as part of complete
recognition systems \cite{posec3d,yan2018stgcn,actorrelations,yun2012}.
Consequently, changes in representation are often accompanied by changes in
encoder capacity, temporal modeling, or supervision, making the representation's
independent contribution difficult to isolate: a richer interaction descriptor
and a more capable model are varied together. As a result, a basic question
remains open---\emph{how much discriminative information does the interaction
representation itself contribute, when everything else is held fixed?}

To answer this question, we design a controlled study. We treat the interaction
representation as the single experimental variable and hold the entire downstream
pipeline constant---the same tracker and pose estimator, the same temporal
detection head, the same weak supervision, folds, normalization, optimizer, and
evaluation. We use early violence detection as the benchmark. This setting is
particularly appropriate because its central challenge is precisely the one
interaction cues are meant to resolve: distinguishing violent close contact from
visually similar benign close contact such as an embrace or a crowd. Under this
controlled protocol we compare five interaction representation variants that span
the spectrum from coarse geometry to a representation learned end to end:
bounding-box geometry, a matched handcrafted pose analogue, two enriched pose
descriptors that add articulation- and contact-specific cues, and a small encoder
learned directly from raw joints.

To evaluate them without the statistical pitfalls common in this small-data
regime, we adopt a video-level protocol: the video, not the frame, is the unit
of analysis; comparisons are paired across shared cross-validation folds; and
uncertainty is estimated with a cluster bootstrap over videos. We report each
comparison under two complementary evaluations---a linear probe that measures
whether the discriminative information is linearly present, and a
multiple-instance detection head that measures whether it survives the
operational pipeline.

Our central finding is consistent across every axis of the study: within this
controlled weakly-supervised early-detection setting, no pose-based
representation---handcrafted or learned---outperforms coarse bounding-box
geometry, and geometry itself is only weakly discriminative. Enriching the pose
descriptor with articulation and contact cues does not yield an improvement whose
confidence interval excludes zero, and a matched-capacity encoder learned from
raw joints also scores below geometry. Because that encoder is evaluated as a
single video-level AUC rather than under the paired probe/MIL protocol, we treat
it as supporting the handcrafted result rather than independently establishing
it. A measured failure analysis over the full anomalous set motivates looking
outside pairwise interaction geometry: the cases that defeat all
representations involve atypical appearance, scene context, and subtle posture
rather than distinctive interaction dynamics.

We emphasize that this is a controlled diagnosis, not a claim about interaction
representations in general. The pose comparison covers a specific family of
handcrafted pairwise descriptors and one deliberately small learned encoder on
UCF-Crime; the appearance, context, and pre-event findings are additionally
tested on XD-Violence. We do not evaluate high-capacity skeleton architectures,
whose additional capacity would reintroduce the very confound our design
removes. Within these bounds the evidence motivates testing whether stronger appearance
and context scores reflect event-specific evidence or information already
available at the video level.

Extending the pipeline beyond interaction cues sharpens this puzzle rather than
resolving it: frozen appearance and full-frame encoders both exceed geometry by
a wide margin, yet cropping to the interacting people never beats encoding the
whole frame. We therefore test what the benchmark measures, and find that most
of the separation is available from frames preceding the annotated event---on
both datasets, and for handcrafted geometry as much as for learned encoders.

\paragraph{Contributions.}
\begin{itemize}
\setlength{\itemsep}{-3pt}
\setlength{\parskip}{0pt}

\item A controlled, capacity-matched comparison of five interaction
representations---geometry, handcrafted pose, enriched pose, and a learned
raw-joint encoder---through one fixed weakly-supervised pipeline, isolating the
representation as the only variable.

\item A video-level protocol with paired cluster-bootstrap confidence intervals
under both a representational (probe) and operational (MIL) test, avoiding
frame-level pseudo-replication; the appearance, context, and pre-event
comparisons are additionally evaluated across four seeds.

\item A two-benchmark ladder showing that representations above coarse geometry
are mutually indistinguishable at UCF-Crime's sample size, while on the nine
times larger XD-Violence split whole-frame context measurably exceeds
person-crop appearance: localizing the interacting people yields no advantage
over encoding the whole frame, and at adequate sample size is worse.

\item A pre-onset diagnostic showing that $39$--$91\%$ of above-chance
separation is available \emph{before} the annotated event on both benchmarks,
under a control removing sequence length as a cue, together with a measured
failure taxonomy and visual identification of concrete provenance artifacts
consistent with this effect.
\end{itemize}

%
%

\section{Related Work}
\label{sec:related}

Interaction and pose representations appear across several research directions
that analyze people in video. Existing work proposes increasingly sophisticated
interaction representations, but the representation itself is rarely isolated as the
experimental variable under a fixed downstream pipeline. We organize prior work into
five groups; each employs interaction or pose information, yet each leaves a
\emph{different} question unanswered about whether richer representations add
discriminative value under matched, weakly-supervised conditions.

\paragraph{Weakly-supervised anomaly and violence detection.}
Under video-level supervision, anomalies are detected via multiple-instance
learning over segment features~\cite{sultani2018,tian2021rtfm}, refined by self-training over pseudo-labelled
snippets~\cite{feng2021mist} or by removing context bias from instance
selection~\cite{lv2023umil}; a related line targets violence on trimmed clip
benchmarks~\cite{rwf2000,rlvs}.
Both optimize complete systems end to end, so the interaction signal is
entangled with the feature encoder and temporal model and the representation's
own contribution is never isolated.

\paragraph{Pose, interaction, and group-activity representations.}
Skeleton methods model actions with graph-convolutional or 3D
encoders~\cite{posec3d,yan2018stgcn}, demonstrating pose's expressive power but
under full supervision and at high capacity. Two-person descriptors and
group-activity models explicitly encode relations between
people~\cite{yun2012,groupactivity,actorrelations}. These propose new encodings
but evaluate them within a chosen model, leaving representation quality
confounded with model capacity---the confound our fixed-head design removes.

\paragraph{Early event anticipation.}
Early-detection methods aim to flag an event before it
completes~\cite{earlyanticipation}, motivating our setting; their focus is
the temporal prediction objective rather than which spatial representation best
supports it.

\paragraph{Our position.}
Despite the widespread use of interaction and pose representations across
surveillance understanding, group-activity recognition, video anomaly detection,
and violence detection, surprisingly little work isolates the interaction
representation \emph{itself} as the object of study. Although all five directions
above employ interaction or pose information, they pursue different goals, so
improvements in one setting do not establish whether richer
interaction representations \emph{themselves} carry more discriminative
information under matched conditions. Our work complements rather than replaces
them. Instead of proposing another detector or another interaction encoder, we
ask a different
question: how much discriminative information is contributed by the interaction
representation \emph{itself} when the downstream learning pipeline is held fixed?
We answer it by varying only the interaction representation under a fixed
weakly-supervised early-detection pipeline, adding video-level statistics and a
measured failure analysis---isolating whether richer interaction representations
carry more discriminative information than coarse geometry in this setting.


%

\section{Study Setup}
\label{sec:setup}

\subsection{Task and data}
We study weakly-supervised early violence detection on
UCF-Crime~\cite{sultani2018}, a standard untrimmed surveillance benchmark
trained with video-level labels; frame-level annotations are used only for
test-time localization and lead-time analysis. Because our question concerns
interaction representations, we focus on the anomalous categories in which close
interaction between people is the defining cue.

\paragraph{Interaction subset selection.}
We restrict the UCF-Crime test set to the categories involving direct
interpersonal interaction: Abuse, Arrest, Assault, and Fighting. The criterion
is applied at the category level rather than per video---every test video in
these four categories is included---yielding $15$ anomalous videos evaluated
with all $150$ normal test videos, $165$ in total. The selection was defined
before experimentation, and we retain every qualifying anomalous video rather
than a curated subset, reviewing all of them in Section~\ref{sec:taxonomy}. The
small anomalous sample is inherent to an interaction-centric subset; we address
it through the video-level protocol of Section~\ref{sec:protocol} and report
uncertainty throughout.

\paragraph{Second benchmark.}
The representation ladder of Section~\ref{sec:ladder} and the diagnostic of
Section~\ref{sec:preonset} are additionally run on
XD-Violence~\cite{wu2020xdviolence}, a larger weakly-supervised benchmark whose
test split also carries frame-level annotations. Applying the same category-level criterion---every test video labelled Fighting
or Abuse, plus all normal test videos---gives $137$ anomalous and $299$ normal
videos, nine times our UCF-Crime anomalous sample. Its non-violent videos were
background-matched to the violent ones to prevent scene-based discrimination,
making it the stronger setting for Section~\ref{sec:preonset}. The
five-representation pose comparison is run on UCF-Crime only.

\subsection{Weakly-supervised detection head}
All representations are evaluated through a single detection head trained under
weak (video-level) multiple-instance supervision: a small causal temporal model
(specified in Section~\ref{sec:head}) whose prediction at any time depends only
on the past, as early detection requires. Each video is resized to a fixed
temporal length so that every optimization step observes full coverage, avoiding
the sampling instability of randomly cropping long untrimmed videos. The head is
shared and held identical across all representations; it is infrastructure for
the comparison rather than a contribution.

\subsection{Baselines and context}
Our objective is not to surpass full-scene anomaly detectors. A strong
weakly-supervised detector (RTFM~\cite{tian2021rtfm}) attains a frame-level AUC
of $0.887$ on this subset against $0.669$ for an interaction-only
detector---establishing that interaction cues alone do not match full-scene
systems, the premise our study investigates rather than a target it aims to
beat. These frame-level values are context only, not comparable to the
video-level AUCs reported here.

\subsection{Early-detection metric}
Detection is scored under a false-alarm-rate--controlled criterion: a video
counts as detected only if the risk score crosses an operating threshold within
the evaluation window, with lead time measured relative to the annotated onset.
The representation comparison summarizes detection by video-level area under the
ROC curve (Section~\ref{sec:protocol}); the lead-time and false-alarm
characterization is used in the failure analysis (Section~\ref{sec:taxonomy}),
where temporal localization becomes a separate axis of interest.


\section{Interaction Representations}
\label{sec:representations}

We compare five pairwise interaction representations spanning a spectrum from
coarse spatial geometry to a representation learned end-to-end from raw joints,
all consumed by an \emph{identical} temporal detection head under identical
training and evaluation (Section~\ref{sec:protocol}). The representation is the
only variable, isolating its discriminative content from confounds of capacity,
optimization, or evaluation.

\subsection{Person tracking and pose estimation}
All representations are built on the same underlying detections. We run a
shared person detector and multi-object tracker (YOLOv8m detections with
BoT-SORT association) at a temporal stride of two, producing consistent track
identities across frames. For the pose-based representations we additionally
run a top-down pose estimator (RTMPose-m, COCO-17 keypoints) on each tracked
box, preserving track identity. Because every representation derives from the
same tracks, differences between them cannot arise from differences in
detection or tracking.

\subsection{Bounding-box geometry (G7)}
Our coarse baseline describes each frame by seven pairwise geometric channels
computed over tracked person boxes: the number of tracked people, the minimum
and mean inter-person center distance, the maximum approach rate (the rate at
which a pair's separation is shrinking), the maximum alignment (how directly one
person moves toward another), the maximum bounding-box overlap (a contact
proxy), and the maximum relative speed. All distances and velocities are
normalized by person height for invariance to camera zoom, and channels are
undefined in frames with fewer than two people. The complete mathematical
definitions of all interaction channels are provided in the supplementary
material. This representation encodes \emph{where bodies are and
how they move} relative to one another, with no information about body
articulation.

\subsection{Handcrafted pose analogue (P7)}
To test whether articulated pose adds discriminative information under a strictly
controlled substitution, we construct a seven-channel pose analogue in
one-to-one correspondence with G7, replacing each box-level quantity with its
skeletal counterpart: min/mean pairwise inter-person joint distance, maximum
joint approach rate, maximum \emph{reach} (wrist extension toward the other
person, the pose analogue of orientation alignment), maximum skeletal-limb
overlap, and maximum joint relative speed. Each channel shares the semantics of
its geometric counterpart, so any difference is attributable to the skeletal
signal rather than to a change in what is measured.

\paragraph{Keypoint standard.}
We key interaction on upper-body and core keypoints (shoulders, elbows, wrists,
hips, nose) rather than a full all-pairs joint descriptor: face and foot
keypoints are unreliable at surveillance resolution, and a high-dimensional
descriptor would confound the controlled comparison against the geometric
channels. Distances are height-normalized as in G7, with a guard discarding
degenerate skeletons (fewer than three confident keypoints) to prevent
normalization instability.

\subsection{Enriched pose (P11 and PR4)}
The one-to-one analogue may understate pose by construction, since it mirrors
box geometry rather than exploiting uniquely articulated cues. We therefore add
four \emph{pose-specific} channels that geometry cannot express (wrist-to-head
and wrist-to-torso distance as strike-target proxies, inter-person elbow-angle
difference, and maximum joint acceleration; defined in the supplement),
evaluated both appended to the tight analogue (\textbf{P11}, eleven channels)
and in isolation (\textbf{PR4}).

\subsection{Learned pose encoder}
Handcrafted representations summarize each skeleton pair into a small set of
scalars, which may discard information a learned representation would retain.
We therefore replace the handcrafted front end with a small encoder trained
end-to-end against the detection objective, holding the temporal head fixed.
For each person pair we form a $104$-dimensional input (both hip-centered,
scale-normalized skeletons, per-joint confidences, and relative torso
displacement) encoded by a shared two-layer MLP ($104\!\to\!32\!\to\!32$,
$\sim\!4$k parameters), applied to both orderings and averaged for permutation
invariance, then max-pooled across pairs. The front end contains no temporal
layers: all temporal modeling remains in the shared head, isolating the effect
of a learned representation from any change in temporal capacity. The encoder
is deliberately small; graph-convolutional or transformer skeleton encoders
would reintroduce the capacity confound this design removes.

\subsection{Shared detection head and control discipline}
\label{sec:head}
All representations feed an identical detection head. It consists of two causal
1D convolution layers (kernel size $5$, dilations $1$ and $2$), followed by a
unidirectional (causal) GRU (hidden size $64$) and a linear sigmoid
classification head---roughly $25$K parameters for the interaction-only
configuration. The head is causal so that a prediction at any time depends only
on past frames, as required for early detection. It is trained under weak
(video-level) supervision with a multiple-instance ranking loss, identical
across all representations, with each video resized to a fixed length so that
every training step observes full video coverage.

The comparison is controlled by construction: across all representations we hold
fixed the videos and splits, the cross-validation folds, the feature
normalization (computed on the training fold only, to prevent leakage), the
validity-mask handling, the fixed-length resizing, the optimizer and its
hyperparameters and seeds, and the evaluation metric. For the four handcrafted
representations the representation is therefore the only experimental variable.
The learned encoder is evaluated as a matched-capacity extension under the same
downstream task, folds, and head, but trains a front end of its own, so its
comparison is not variable-isolated in the same strict sense.


\section{Evaluation Protocol}
\label{sec:protocol}

\subsection{The video as the unit of analysis}
A frame-level analysis that pools interaction frames across videos treats
thousands of highly correlated frames as independent observations, inflating
effective sample size and understating uncertainty---a form of
pseudo-replication. We therefore treat the \emph{video} as the unit of
analysis throughout. Any frame-level separability statistic we report is
descriptive only; all inferential comparisons are computed at the video level
with uncertainty estimated by resampling whole videos.

\subsection{Two complementary tests}
We evaluate each representation two ways. The \textbf{linear probe} is a purely
representational test: for each video we form a fixed-length signature from
robust summary statistics of its interaction channels, and train a logistic
regression classifier under video-disjoint cross-validation. Because the probe
has no temporal capacity of its own, it measures whether the discriminative
information is \emph{linearly present} in the representation, with no temporal
model able to mask or manufacture separability. The \textbf{MIL evaluation} is
the operational test: the full detection head is trained on each representation
under the same folds, measuring whether the representation supports detection
in the actual pipeline. Reporting both guards against a result that is an
artifact of either the linear simplification or the neural training procedure.

\subsection{Repeated cross-validation and cluster bootstrap}
Both tests use repeated stratified $k$-fold cross-validation ($k=5$, ten
repeats), with the fold partitions \emph{shared across all representations} so
that per-video predictions are directly paired. From the out-of-fold
predictions we compute, for each representation, an aggregate video-level AUC.
Uncertainty on the difference between two representations is estimated by a
\textbf{cluster bootstrap} that resamples entire videos (2000 resamples);
because the same resampled video set is scored under both representations, this
yields a paired $\Delta$AUC and its $95\%$ confidence interval. We report
$\Delta$AUC and its interval for the comparisons of interest---each pose
representation against geometry, and the enriched pose against the tight
analogue---and consider a difference meaningful only when its interval excludes
zero.

\paragraph{Interpretation criterion.}
To avoid post-hoc selection among noisy comparisons, we specified the
interpretation criterion before running the final evaluation: enriched pose is
considered to carry additional discriminative information only if it exceeds
\emph{both} geometry and the tight pose analogue with a paired-difference
interval excluding zero, in the same direction under both the probe and the MIL
test. A single favorable fold or a single favorable test is not treated as
evidence.

\subsection{Matched video set}
All representations are evaluated on the same matched set of $165$ videos ($15$
anomalous and $150$ normal). Each pairwise representation is standardized
per-channel; frames with no valid tracked person-pair (whose feature values are
undefined) are set to zero after standardization, and a separate binary validity
channel is appended marking which frames were finite. For the seven-channel
geometry this yields eight input channels ($7$ features $+$ $1$ validity mask),
so the model is explicitly told which frames carry no interaction rather than
being misled by a zero that could be mistaken for a real measurement. Thirty
videos contain no valid tracked person-pair in any frame; these are retained
(not dropped selectively) and enter the model as all-zero feature sequences with
the validity channel off throughout, under the same convention as every other
representation---keeping the sample and fold assignments identical across
comparisons.


\section{Results}
\label{sec:results}

We report the video-level comparison of the five interaction representations
under both evaluation paradigms of Section~\ref{sec:protocol}. All values are
computed on the matched set of $165$ videos ($15$ anomalous), using shared
cross-validation folds and the cluster bootstrap over videos.

\begin{table*}[t]
\centering
\footnotesize
\renewcommand{\arraystretch}{0.95}
\setlength{\tabcolsep}{5pt}
\begin{tabular}{lccccc}
\toprule
& \multicolumn{2}{c}{AUC} & & \multicolumn{2}{c}{Paired $\Delta$AUC vs.\ G7 $[95\%$ CI$]$} \\
\cmidrule(lr){2-3} \cmidrule(lr){5-6}
Representation & Probe & MIL & & Probe & MIL \\
\midrule
G7 \ (geometry, 7)        & $\mathbf{0.724}$ & $\mathbf{0.684}$ & & --- & --- \\
P7 \ (tight pose, 7)      & $0.580$ & $0.604$ & & $-0.153\,[\text{-}0.317,\text{-}0.011]^{*}$ & $-0.080\,[\text{-}0.201,0.043]$ \\
P11 (tight+rich, 11)      & $0.624$ & $0.632$ & & $-0.104\,[\text{-}0.259,0.034]$ & $-0.049\,[\text{-}0.209,0.109]$ \\
PR4 (rich only, 4)        & $0.643$ & $0.597$ & & $-0.081\,[\text{-}0.222,0.039]$ & $-0.084\,[\text{-}0.233,0.068]$ \\
Learned pose (raw joints)$^{\dagger}$ & \multicolumn{2}{c}{$0.631$} & & --- & --- \\
\midrule
\multicolumn{6}{l}{\emph{Pose-over-pose:}\ \ P11 $-$ P7\ \ probe $+0.047\,[\text{-}0.001,0.104]$;\ \ MIL $+0.034\,[\text{-}0.118,0.173]$}\\
\bottomrule
\end{tabular}
\caption{Video-level AUC for each interaction representation under a linear
probe (representational) and MIL (operational) evaluation, with paired
$\Delta$AUC and $95\%$ cluster-bootstrap confidence intervals over videos.
Folds are shared across all representations. $^{*}$ marks the only paired
difference whose interval excludes zero. $^{\dagger}$The learned encoder is
trained end-to-end against the detection objective and evaluated as a single
video-level AUC under the same repeated cross-validation; it does not decompose
into a separate probe/MIL pair, so we do not compute a paired interval against
G7 for it. Because of this asymmetry, statements that both evaluation paradigms
agree refer to the four handcrafted representations; the learned encoder is
reported separately and also falls below geometry.}
\label{tab:main}
\end{table*}

\begin{figure}[tb]
\centering
\includegraphics[width=0.88\linewidth]{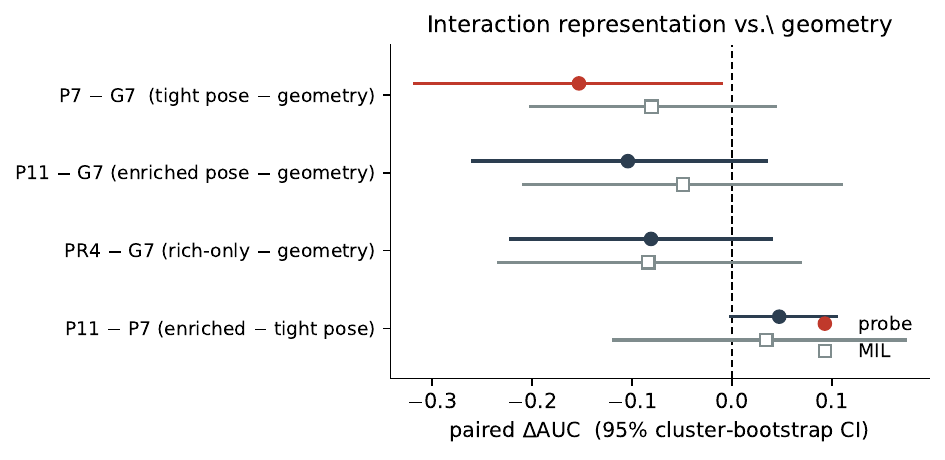}
\caption{Paired $\Delta$AUC (representation minus geometry) with $95\%$
cluster-bootstrap confidence intervals, for the linear probe and MIL
evaluations. Intervals crossing the dashed zero line indicate no difference
from coarse geometry that survives video-level uncertainty; only the tight pose
analogue under the probe (P7$-$G7) falls entirely below zero.}
\label{fig:delta}
\end{figure}

\subsection{Main comparison}
Table~\ref{tab:main} and Figure~\ref{fig:delta} summarize the comparison. Under
the linear probe, coarse geometry (G7) attains the highest AUC ($0.724$) of the
five representations. The tight pose analogue (P7) is significantly worse
(P7$-$G7 $=-0.153$, $[-0.317,-0.011]$)---the only comparison whose interval
excludes zero. The enriched representations P11 and PR4 trail geometry with
intervals that include zero, and adding pose-specific channels to the tight
analogue (P11$-$P7 $=+0.047$, $[-0.001,0.104]$) does not reach significance. The
learned encoder attains $0.631$, within the handcrafted-pose range
($0.580$--$0.643$) and below geometry, so it too did not improve discrimination
over the handcrafted descriptors or coarse geometry.

\subsection{Agreement across paradigms}
For the four handcrafted representations the relative ordering is consistent
across both paradigms: every paired comparison against geometry in
Table~\ref{tab:main} is negative in central value under both probe and MIL, and
the one pose-over-pose comparison (P11$-$P7) is positive but not significant
under either. The separately evaluated learned encoder also remains below
geometry ($0.631$). Agreement across two methodologically distinct evaluations
indicates the pattern is not an artifact of either the linear simplification or
the neural training procedure. A feature-space visualization of per-video pose
signatures, in which anomalous videos are dispersed among normal videos rather
than forming a separated cluster, is provided in the supplement.

\subsection{Summary of observations}
In short, no pose-based representation---handcrafted or learned---exceeded coarse
geometry, the tight analogue was significantly worse under the probe, and probe
and MIL agreed in direction throughout. Coarse geometry itself reached only
$0.724$ (probe) / $0.684$ (MIL), which---with the situations where all
representations fail---we examine in Section~\ref{sec:taxonomy}.

\subsection{Beyond interaction: appearance and scene context}
\label{sec:ladder}

Our failure analysis points outside pairwise geometry, so we extend the same
pipeline with \textbf{APP}, the union region of the tracked people encoded by a
frozen DINOv2 ViT-B/14~\cite{oquab2024dinov2}, and \textbf{CTX}, the full frame
encoded by the same model. Both feed the identical head on identical folds under
the identical missing-frame convention. We also evaluate raw-concatenation
fusions, and rerun the ladder on XD-Violence.

\begin{table}[t]
\centering
\footnotesize
\renewcommand{\arraystretch}{0.95}
\setlength{\tabcolsep}{3pt}
\begin{tabular}{lcccc}
\toprule
& \multicolumn{2}{c}{UCF ($n\!=\!165$)}
& \multicolumn{2}{c}{XD ($n\!=\!436$)} \\
\cmidrule(lr){2-3}\cmidrule(lr){4-5}
Rep. & AUC & vs.\ G7 & AUC & vs.\ G7 \\
\midrule
G7                 & $.685_{\pm.024}$ & ---              & $.829_{\pm.009}$ & --- \\
APP                & $.892_{\pm.016}$ & $+.207^{4/4}$ & $.924_{\pm.007}$ & $+.095^{4/4}$ \\
CTX                & $.907_{\pm.016}$ & $+.219^{4/4}$ & $.959_{\pm.002}$ & $+.130^{4/4}$ \\
\bottomrule
\end{tabular}
\caption{The MIL representation ladder on both benchmarks, each under its own
folds and head. Mean$_{\pm\text{sd}}$ over four seeds; the superscript gives the
number of seeds whose paired cluster-bootstrap interval excludes zero. G7 =
geometry, APP = person crops, CTX = full frame. Coarse geometry is worse than
every frozen-encoder representation on both datasets, in every seed.}
\label{tab:ladder}
\end{table}

Table~\ref{tab:ladder} reports the result. Coarse geometry is the clear floor:
every representation above it improves by $+0.09$ to $+0.22$ AUC, intervals
excluding zero in all four seeds on both benchmarks---the largest and most
stable effect in the study.

The comparisons \emph{among} these representations are more informative, and the
datasets disagree instructively. On UCF-Crime none differs from any other: all
fifteen pairwise comparisons among APP, CTX, and the fusions flip sign across
seeds and exclude zero in at most one of four. On XD-Violence it resolves: CTX exceeds APP by
$0.028$--$0.048$ AUC under MIL and $\approx\!0.047$ under the probe, intervals
excluding zero in $4/4$ seeds under both. We attribute this to statistical power: $15$ anomalous videos cannot resolve an
effect of about $0.04$ AUC and $137$ can. The larger benchmark therefore
supports a clearer conclusion: person cropping provides no advantage over
full-frame context, and on that split is measurably \emph{worse} than it.

One negative result illustrates why the four-seed protocol matters: a
G7\,+\,APP gain over APP whose interval excluded zero at a single UCF-Crime
seed reversed sign across the others, and we withdraw it (supplement).

That a full-frame embedding, given no information about which people are
interacting, matches or exceeds every interaction-aware representation questions
whether the measured AUC primarily reflects interaction-specific evidence.
Section~\ref{sec:preonset} tests this directly.


\section{What are these AUCs measuring?}
\label{sec:preonset}

Section~\ref{sec:ladder} raises an important confounding possibility. If a
full-frame encoder with no notion of people matches every interaction-aware
representation, the shared signal may not be the violent event: the anomalous
and normal test videos are not drawn from a common source, so a representation
encoding camera placement, venue, framing, or image quality separates the
classes without observing an event. Benchmarks are well known to admit such
shortcuts~\cite{geirhos2020shortcut}, and provenance can be recoverable from
images alone~\cite{torralba2011bias}; we test whether that occurs here, and how
much of the reported separation it accounts for.

\paragraph{A direct test.}
Each anomalous video is truncated to the frames \emph{before} its annotated
onset while keeping its label. Both arms use the same videos, folds, head, and
seed, so the only difference is which frames the anomalous videos contribute.
We report

\vspace{-2pt}
\[
\text{retained} \;=\; \frac{\mathrm{AUC}_{\text{pre}} - 0.5}
                            {\mathrm{AUC}_{\text{full}} - 0.5},
\]

\vspace{-2pt}
\noindent the fraction of above-chance separation still available before the
event: near $0$ if a representation relies on event evidence, near $1$ if it
identifies videos. We run the test on both benchmarks; XD-Violence is the stronger of the two,
since its non-violent videos were background-matched to the violent ones to
prevent scene-based discrimination. Truncation shortens anomalous videos, so we
report as primary a length-matched arm in which each normal video is truncated
to a length drawn from the anomalous pre-onset distribution; this reduces
XD-Violence context retention from $98\%$ to $70\%$, and the effect survives.

\begin{table}[t]
\centering
\scriptsize
\renewcommand{\arraystretch}{0.95}
\setlength{\tabcolsep}{4pt}
\begin{tabular}{lcccc}
\toprule
& \multicolumn{2}{c}{UCF-Crime} & \multicolumn{2}{c}{XD-Violence} \\
\cmidrule(lr){2-3}\cmidrule(lr){4-5}
Representation & Probe & MIL & Probe & MIL \\
\midrule
G7 (geometry)      & $79\%$ & $84\%$ & $46\%$ & $39\%$ \\
APP (person crops) & $82\%$ & $70\%$ & $57\%$ & $50\%$ \\
CTX (full frame)   & $91\%$ & $83\%$ & $70\%$ & $71\%$ \\
\bottomrule
\end{tabular}
\caption{Separation retained before the event, under the length-matched control
(normal videos truncated to lengths drawn from the anomalous pre-onset
distribution, so sequence length carries no class information). Entries are
$(\mathrm{AUC}_{\text{pre}}-0.5)/(\mathrm{AUC}_{\text{full}}-0.5)$, the
fraction of above-chance separation available from pre-onset frames only; means
over four seeds. Every representation on both benchmarks retains a substantial
fraction. Full-video AUCs and the unmatched arm are tabulated per seed in the
supplement.}
\label{tab:preonset}
\end{table}

\begin{figure}[t]
\centering
\includegraphics[width=0.82\linewidth]{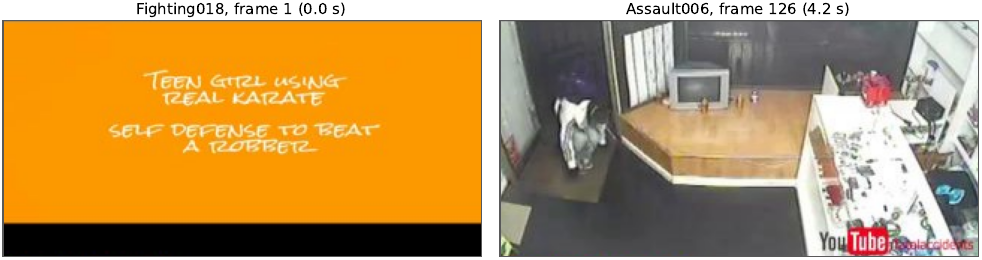}
\caption{Two pre-onset frames from UCF-Crime anomalous videos---footage the
annotation marks as preceding the event. Left: the Fighting018 window contains
no footage, only an editorial title card naming the video's content. Right: a
platform watermark is composited into every frame of Assault006, whose
$39$-second window is free of violence. Neither occurs in the surveillance
footage supplying the normal class: both are class-informative without being
event-informative---concrete instances of a source-distribution mechanism, not
a complete account of it.}
\label{fig:artifacts}
\end{figure}

\paragraph{Result.}
Table~\ref{tab:preonset} shows that a substantial fraction of the measured
separation is available before the event, for every representation on both
benchmarks: probe retention is $79$--$91\%$ on UCF-Crime and $46$--$70\%$ on
XD-Violence. It is uniformly lower on XD-Violence, consistent with its
background-matched design reducing---though not removing---source cues, and the
representations separate there, context retaining most and geometry least. What
holds throughout is that seven hand-designed channels, encoding only how many
people are present and how they are arranged, still recover $39$--$84\%$ of
above-chance separation from footage in which nothing has yet happened.

\paragraph{Are the onset labels simply late?}
We inspected the pre-onset windows of the five videos with the tightest margins
on each dataset, where a late label would contaminate the window most easily.
On XD-Violence all five were free of visible violence; on UCF-Crime three of
five were, and in one the altercation is underway. Two cases are instructive rather than problematic
(Figure~\ref{fig:artifacts}): an editorial title card occupying one video's
entire pre-onset window, and a platform watermark in every frame of another.
Neither depicts an event, nor appears in the surveillance footage supplying the
normal class. We do not claim such artifacts account for the whole effect---videos may also
differ in viewpoint, resolution, compression, and scene type---but they show
that class-informative, event-uninformative cues exist.

\paragraph{Pre-onset separability is not synonymous with leakage.}
Some pre-event separability is legitimate: escalation and approach are exactly
what an early detector should exploit, and the diagnostic cannot by itself
separate anticipation from source cues. Three observations nonetheless indicate that much of it
is not event evidence: retention is comparable for hand-designed channels and
frozen encoders, so it is not large models reading subtle precursors; inspected
windows are mostly free of any visible precursor; and the artifacts of
Figure~\ref{fig:artifacts} carry no temporal information at all, yet are
perfectly class-informative.

\paragraph{What this implies.}
We do not conclude that these benchmarks are uninformative, nor that prior
results on them are wrong, but something narrower: a video-level AUC here is a composite of event evidence and pre-event source
cues, so comparisons \emph{between} representations are ones in which much of
the measured quantity is shared and not attributable to the event.


\section{Failure Analysis}
\label{sec:taxonomy}

To characterize \emph{where} discrimination breaks down, we review all fifteen
interaction-category anomalous UCF-Crime test videos and compute a failure
signature for each. Three patterns are measurable, defined with per-video values
in the supplement: \textbf{C1}, absent interaction (two
videos have zero tracked person-pairs, so the pairwise stream fails); \textbf{C2}, atypical appearance (three videos have
an I3D centroid distance of $11.9$--$12.6$ against $\approx\!3.0$ elsewhere);
and \textbf{C3}, quiet violence, a residual group missed despite ordinary
appearance distance and nonzero person count.

Across four seeds, four videos are never robustly detected by \emph{any}
representation, including the frozen encoders of Section~\ref{sec:ladder}: both
C1 cases and two of three C2 cases. Context recovers none, though it is defined
on every frame of the C1 videos where the other streams are not. These are not
failures a richer pairwise descriptor would remedy.

\section{Discussion}
\label{sec:discussion}

Because only the representation varied across the handcrafted comparison,
differences there are attributable to representational content rather than model
capacity or training.

\paragraph{What the evidence shows.}
No pose representation exceeded coarse geometry under either test
(Table~\ref{tab:main}). Extending the ladder to frozen encoders on both
benchmarks (Table~\ref{tab:ladder}) sharpened rather than resolved the picture:
appearance and full-frame context both exceed geometry by a wide margin, yet on
UCF-Crime they are indistinguishable from one another and from every fusion,
while on the larger XD-Violence split whole-frame context measurably exceeds
person-crop appearance in every seed. Localizing the interacting people
therefore yields no advantage over encoding the whole frame.

\paragraph{Reinterpreting the comparison.}
Section~\ref{sec:preonset} changes how these numbers should be read. If $39$--$91\%$ of the measured separation is available \emph{before} the
annotated event, on both datasets and for hand-designed geometry as well as
frozen encoders, then a video-level AUC here is not a measure of event detection
alone, and differences between representations are differences in a quantity
substantially shared and not attributable to the event. We therefore do not
conclude that pose carries no interaction information, but that this evaluation
cannot detect it---a statement about the benchmark rather than about pose. Context bias here has been addressed by changing the
learner~\cite{lv2023umil}; we suggest also measuring it directly. For
deployment, high AUC from pre-event cues may overstate event recognition.

\paragraph{Limitations.}
The five-representation pose comparison exists only on UCF-Crime, whose
anomalous subset is small (fifteen videos), so its intervals are wide---and we
show the consequence, since the appearance--context difference indistinguishable
at that sample size resolves clearly on XD-Violence's $137$. The comparison
fixes model capacity by design, leaving open whether larger skeleton-specific
architectures recover additional signal. The diagnostic inherits the benchmarks'
annotation timing: of ten tight-margin windows inspected, one had a genuinely
late onset. Finally, retention differs markedly between benchmarks
($46$--$70\%$ against $79$--$91\%$ under the probe), so its magnitude is
dataset-dependent even though its presence is not. Applying the diagnostic to further benchmarks, and isolating which cues drive
the retained separation, are the natural next steps.

\section{Conclusion}
\label{sec:conclusion}

Varying only the interaction representation, no pose descriptor beat coarse
geometry, while frozen encoders improved discrimination without needing to
localize the interacting pair; on the larger XD-Violence split, whole-frame
context exceeded person-crop appearance. A pre-onset test shows why these
comparisons require caution: on both benchmarks substantial separation remains
before the annotated event, including for handcrafted geometry, and those
windows contain provenance artifacts no normal video carries. We read this as a
limit on what video-level AUC establishes about representation quality, not as
evidence that pose is uninformative. Before asking which representation is
best, we must ask how much of it is the event.

{
    \small
    \bibliographystyle{ieeenat_fullname}
    \bibliography{main}
}

\end{document}